\documentclass[11pt,a4paper]{article}

\usepackage[a4paper,margin=2.35cm]{geometry}
\usepackage[T1]{fontenc}
\usepackage[utf8]{inputenc}
\usepackage{lmodern}
\usepackage{amsmath,amssymb,amsthm,mathtools,bm}
\usepackage{booktabs,array,longtable,multirow}
\usepackage{enumitem}
\usepackage{hyperref}
\usepackage{xcolor}
\usepackage{graphicx}
\usepackage{pgfplots,pgfplotstable}
\usepgfplotslibrary{groupplots}
\pgfplotsset{compat=1.18}
\usepackage{caption}
\usepackage{subcaption}
\usepackage{float}
\usepackage{microtype}
\usepackage[normalem]{ulem}
\usepackage{listings}
\usepackage{algorithm}
\usepackage{algpseudocode}

\hypersetup{colorlinks=true,linkcolor=blue,citecolor=blue,urlcolor=blue}
\setlist[itemize]{leftmargin=2em}
\setlist[enumerate]{leftmargin=2.2em}
\allowdisplaybreaks
\newtheorem{definition}{Definition}[section]
\newtheorem{theorem}[definition]{Theorem}
\newtheorem{proposition}[definition]{Proposition}
\newtheorem{corollary}[definition]{Corollary}
\theoremstyle{remark}

\newcommand{\E}{\mathbb{E}}

\newcommand{\I}{\mathbb{I}}

\newcommand{\Binom}{\mathrm{Binom}}

\newcommand{\Tr}{\mathrm{tr}}

\newcommand{\ARMS}{\mathrm{ARMS}}
\newcommand{\Var}{\mathrm{Var}}
\newcommand{\Cov}{\mathrm{Cov}}

\title{\textbf{RL2ML: Finite-Rollout Surrogate Objectives from Reinforcement Learning to Maximum Likelihood}}

\author{
  \textbf{Yifu Zheng}\\[1mm]
  University of Southern California\\[1mm]
  \texttt{henryzhe@usc.edu}
}

\date{}

\begin{document}
\maketitle

\begin{abstract}
Correctness-based Reinforcement Learning with Verifiable Rewards (RLVR) trains language models from binary feedback on sampled outputs, but the objective optimized in expectation and the stochastic update geometry induced by finite rollout groups are often conflated. This paper develops RL2ML, a family of finite-rollout surrogate objectives with a closed-form, exactly unbiased gradient estimator. The family continuously connects standard reinforcement learning, maximum-likelihood-like training, and beyond-maximum-likelihood objectives while preserving estimator-objective alignment under a fixed rollout budget. We introduce the group-level update scale to characterize how a rollout group is reweighted after its empirical success count is observed, revealing a subcritical-supercritical update-scale transition that is hidden by population-level objective notation alone. Building on this distinction, calibrated metric-gain analysis and exact variance decomposition show that the best choice of surrogate objective is determined neither by proximity to maximum likelihood nor by the population-level weight alone. Instead, it depends jointly on the evaluation metric, local sensitivity, and estimator variance. The remaining degree of freedom in the surrogate objective family can therefore be formulated as a one-dimensional optimization problem rather than treated as an unconstrained hyperparameter.
\end{abstract}

\section{Introduction}
For a learning task with binary verifiable feedback, let $p:=p_\theta(x)$ denote the probability that a model with parameter $\theta$ produces a correct output for prompt $x$. In fully differentiable supervised learning, maximizing log-likelihood objectives is a standard statistical principle and gives rise to classical losses such as those used in linear regression and logistic regression\cite{murphy2012}. In the single-prompt abstraction considered here, the maximum-likelihood objective is $\mathcal{J}^{\mathrm{ML}}_\theta(x)=\log p$.

RLVR can also be interpreted as changing the probability mass assigned to response trajectories, but it usually operates through sampled rollouts and binary verifier rewards rather than direct likelihood labels. This naturally raises the question studied by MaxRL\cite{maxrl}, which shows that correctness-based RLVR optimizes only a first-order approximation to maximum likelihood and derives a compute-indexed surrogate objective by expanding maximum-likelihood objective. In training, the MaxRL objective gives larger weight to low-success prompts than ordinary reinforcement learning in a maximum-likelihood-like form.

However, even though the MaxRL objective approaches the maximum-likelihood objective in the large-rollout limit, this does not automatically imply that it is the best objective for finite-horizon RLVR training. In a more realistic RL training setting, under finite rollout budgets, finite optimization horizons, and concrete validation metrics, we must ask how to allocate weight to low-success prompts more effectively, and how to design a surrogate objective on that basis.

This is precisely the problem this paper attempts to address. This paper introduces RL2ML, a finite-rollout surrogate objective family that preserves the estimator-objective alignment of MaxRL while exposing a single continuous degree of freedom. For $\gamma\ge 0$, the untruncated power-likelihood gradient is
\begin{align*}
\nabla_\theta \mathcal{J}^{\mathrm{RL2ML}}_\gamma(x)
=
\bigl(p_\theta(x)\bigr)^{-\gamma}\nabla_\theta p_\theta(x).
\end{align*}
The point $\gamma=0$ recovers the ordinary reinforcement learning gradient, while $\gamma=1$ recovers the maximum-likelihood gradient. $\gamma < 1$ form a subcritical regime relative to the ML boundary, whereas values above one form a supercritical regime that upweights low-success prompts beyond maximum likelihood. Along this path, RL2ML continuously connects ordinary RL, maximum-likelihood-style training, and beyond-maximum-likelihood low-success amplification.

The contributions are fourfold. First, RL2ML defines a truncated power-likelihood surrogate objective and provides closed-form coefficients for an exactly unbiased estimator under a finite rollout budget. Second, the group-level update scale $\alpha_K$ exposes a subcritical-supercritical boundary at $\gamma=1$, clarifying how empirical success counts change the actual stochastic update. Third, calibrated local-gain analysis and exact variance decomposition show why supercritical weighting of the hardest prompts need not dominate in finite-horizon training. Fourth, the choice of $\gamma$ is formulated as a one-dimensional outer optimization problem, under which $\gamma=1$ is only an important reference point and a structural boundary, not a universal optimum.

The rest of the paper is organized as follows. Section~\ref{sec:preliminaries} introduces the setup, the MaxRL estimator, and the Bernstein representation of $K$-only estimators. Section~\ref{sec:rl2ml} defines the RL2ML surrogate objective, derives the unbiased estimator, and analyzes its update-scale geometry. Section~\ref{sec:finite-horizon-selection} develops the calibrated local-gain criterion and the variance-aware outer selection rule. Section~\ref{sec:related-work} positions RL2ML relative to prior induced-objective, policy-gradient, and RLVR work, and Section~\ref{sec:conclusion} concludes. Appendix~\ref{app:proofs} gives proofs and derivations, Appendix~\ref{app:verl} gives implementation details for fixed-$\gamma$ estimation and $\gamma^*$ selection, and Appendix~\ref{app:triad} records the triad-family extension.

\section{Preliminaries}
\label{sec:preliminaries}

\subsection{Basic Definitions}
Let the input space be $\mathcal X$, the latent rollout space be $\mathcal Z$, and the task distribution be $x\sim\rho$. For a given prompt $x$, the model first samples a latent rollout $z\sim m_\theta(\cdot\mid x)$ and then obtains the final answer through a deterministic decoding function $y=f(z)$. Denote the correct answer by $y^*(x)$. The corresponding correctness reward is defined as $r(x,z):=\I{f(z)=y^*(x)}$, and the single-sample success probability is therefore $p_\theta(x):=\E_{z\sim m_\theta(\cdot\mid x)}[r(x,z)]$. For a fixed prompt $x$, we write $p:=p_\theta(x)$ for brevity.

Let the score function be
\begin{align*}
S(x,z):=\nabla_\theta\log m_\theta(z\mid x).
\end{align*}
This score-function form is the basis of REINFORCE and policy-gradient methods\cite{williams1992reinforce,sutton1999policygradient}.
When $N$ rollouts are sampled independently, write $r_i:=r(x,z_i)$, $S_i:=S(x,z_i)$, and $K:=\sum_{i=1}^N r_i$. If $K\ge 1$, define the average score over successful samples by
\begin{align*}
\bar S_K(x)
:=
\frac1K\sum_{i=1}^N r_iS_i
=
\frac1K\sum_{i=1}^N r(x,z_i)\nabla_\theta\log m_\theta(z_i\mid x).
\end{align*}
We abbreviate $\bar S_K(x)$ as $\bar S_K$. Define the success-conditioned first and second moments by
\begin{align*}
\mu_x:=\E[S\mid r=1,x],\ \text{and}\ \Sigma_x:=\Cov(S\mid r=1,x).
\end{align*}
Since $\nabla_\theta\mathcal{J}^{\mathrm{ML}}(x)=\E_{z\sim m_\theta(\cdot\mid x)}[\nabla_\theta\log m_\theta(z\mid x)\mid f(z)=y^*(x)]$ by Theorem 1 of the MaxRL paper \cite{maxrl},
\begin{align*}
\mu_x=\nabla_\theta\log p_\theta(x),\ \text{and}\ \nabla_\theta p_\theta(x)=p_\theta(x)\mu_x.
\end{align*}
Thus, in correctness-based latent-generation models, the average score function of successful trajectories estimates the maximum-likelihood gradient direction.

\subsection{MaxRL Estimator}
For a single prompt $x$, the population gradients of ordinary reinforcement learning and maximum likelihood are
\begin{align*}
\nabla_\theta\mathcal{J}^{\mathrm{RL}}(x)&=\nabla_\theta p,\\
\nabla_\theta\mathcal{J}^{\mathrm{ML}}(x)&=\nabla_\theta\log p=\frac1p\nabla_\theta p.
\end{align*}
The corresponding population-level weights are $w_{\mathrm{RL}}(p)=1$ and $w_{\mathrm{ML}}(p)=1/p$. MaxRL uses the expansion
\begin{align*}
\log p=-\sum_{k=1}^\infty\frac{(1-p)^k}{k},
\end{align*}
which implies
\begin{align*}
\nabla_\theta\log p
=
\sum_{k=1}^\infty\frac1k\nabla_\theta\bigl(1-(1-p)^k\bigr)
=
\sum_{k=1}^\infty\frac1k\nabla_\theta\pass@k(x).
\end{align*}
The truncated MaxRL objective is therefore
\begin{align*}
\mathcal{J}^{\mathrm{MaxRL}}_T(x)
=
\sum_{k=1}^T\frac1k\pass@k(x).
\end{align*}
MaxRL uses the estimator
\begin{align*}
\hat g_N^{\mathrm{MaxRL}}(x)
=
\begin{cases}
\dfrac1K\sum_{i=1}^N r_iS_i, & \text{if }K\ge 1,\\
0, & \text{if }K=0.
\end{cases}
\end{align*}
It is unbiased for the truncated MaxRL objective with $T=N$. This exact alignment between a finite-rollout objective and a finite-rollout gradient estimator is the structural property preserved by RL2ML.

\subsection{Bernstein Representation of \texorpdfstring{$K$}{K}-only Estimators}
The closed-form estimator of RL2ML relies on a general fact: any estimator whose scalar coefficient depends on a rollout group only through the success count $K$ realizes a Bernstein-polynomial population weight. Related Bernstein-type representations for binary-reward reasoning RL are also discussed in \cite{objective}.

\begin{theorem}[Bernstein representation of $K$-only estimators]
\label{thm:bernstein}
Consider the estimator
\begin{align*}
\hat g_f(x)=f(K)\sum_{i=1}^N r_iS_i,\ \text{where}\ K=\sum_{i=1}^N r_i.
\end{align*}
Let
\begin{align*}
\beta_m:=Nf(m+1),\ \text{for}\ m=0,\ldots,N-1,
\end{align*}
and denote the Bernstein basis by
\begin{align*}
B_{m,N-1}(p):=\binom{N-1}{m}p^m(1-p)^{N-1-m}.
\end{align*}
Then, for any fixed prompt $x$,
\begin{align*}
\E[\hat g_f(x)\mid x]
=
\left(\sum_{m=0}^{N-1}\beta_mB_{m,N-1}(p)\right)\nabla_\theta p
=
w_f(p)\nabla_\theta\mathcal{J}^{\mathrm{RL}}(x),
\end{align*}
where $w_f(p)=\sum_{m=0}^{N-1}\beta_mB_{m,N-1}(p)$ is the population-level weighting function.

Hence the population-level weight of any $K$-only estimator is a Bernstein polynomial of degree $N-1$.
\end{theorem}

The proof is given in Appendix~\ref{app:proof-thm-bernstein}. The theorem separates two notions that are often implicitly merged. The population-level weight $w_f(p)$ specifies the expected gradient of the surrogate objective, whereas the finite-rollout update observed by the optimizer is governed by the empirical success count. Define the group-level update scale
\begin{align*}
\alpha_K:=\frac{K}{N}\beta_{K-1},\ \text{for}\ K\ge 1.
\end{align*}
Then, conditional on $K\ge 1$,
\begin{align*}
\hat g_f(x)
=
\frac{\beta_{K-1}}{N}\sum_{i=1}^N r_iS_i
=
\alpha_K\bar S_K.
\end{align*}
The direction $\bar S_K$ is the average score function of successful trajectories; the scalar $\alpha_K$ determines how strongly that direction is attenuated or amplified after the rollout group is observed. Moreover, conditional on $K=k\ge 1$,
\begin{align*}
\E[\hat g_f(x)\mid x,K=k]&=\alpha_k\mu_x,\\
\Cov(\hat g_f(x)\mid x,K=k)&=\frac{\alpha_k^2}{k}\Sigma_x.
\end{align*}
Thus $\alpha_K$ determines the sample-level update geometry more directly than the population weight $w_f(p)$ alone.

\section{RL2ML}
\label{sec:rl2ml}

\subsection{Single-Prompt RL2ML Surrogate Objective}
\subsubsection{RL2ML Objective Family}
Define the RL2ML objective family by
\begin{align*}
\mathcal{J}^{\mathrm{RL2ML}}_\gamma(x)
=
\phi_\gamma(p_\theta(x)),\ \text{where}\ \phi_\gamma'(p)=p^{-\gamma}.
\end{align*}
We choose the antiderivative as
\begin{align*}
	\phi_\gamma(p) = \begin{cases} \dfrac{p^{1-\gamma}-1}{1-\gamma}, & \gamma\neq 1,\\[0.8em] \log p, & \gamma=1. \end{cases}
\end{align*}
Consequently, $\gamma=0$ recovers the ordinary RL population gradient and $\gamma=1$ recovers the ML population gradient. The untruncated population gradient is
\begin{align*}
\nabla_\theta\mathcal{J}^{\mathrm{RL2ML}}_\gamma(x)
=
\bigl(p_\theta(x)\bigr)^{-\gamma}\nabla_\theta p_\theta(x).
\end{align*}
Finite-rollout training does not estimate this untruncated target directly. Following the MaxRL principle, RL2ML instead defines a rollout-budget-aligned truncated surrogate objective.

\subsubsection{Truncated RL2ML Objective}
For rollout budget $N$, define
\begin{align*}
\mathcal{J}^{\mathrm{RL2ML}}_{\gamma,N}(x)
:=
\sum_{k=1}^N
\frac{(\gamma)_{k-1}}{(k-1)!\,k}\pass@k(x),
\end{align*}
where $(\gamma)_{k-1}$ is the rising factorial and $(\gamma)_0=1$. Its associated population-level weight is
\begin{align*}
w^{\mathrm{RL2ML}}_{\gamma,N}(p)
:=
\sum_{m=0}^{N-1}\frac{(\gamma)_m}{m!}(1-p)^m.
\end{align*}

\begin{proposition}[Gradient of the truncated RL2ML objective]
\label{prop:gradexp}
For every fixed prompt $x$,
\begin{align*}
\nabla_\theta\mathcal{J}^{\mathrm{RL2ML}}_{\gamma,N}(x)
=
w^{\mathrm{RL2ML}}_{\gamma,N}(p_\theta(x))\nabla_\theta p_\theta(x).
\end{align*}
\end{proposition}

The proof is given in Appendix~\ref{app:proof-prop-gradexp}. When $N$ is large, $w^{\mathrm{RL2ML}}_{\gamma,N}(p)$ approaches the power weight $p^{-\gamma}$ on the interior of $(0,1]$. For finite $N$, it is a truncated, rollout-budget-aligned approximation whose gradient admits an exactly unbiased estimator.

\subsection{Exactly Unbiased Gradient Estimator}
\subsubsection{Closed-Form Coefficients}
By Theorem~\ref{thm:bernstein}, it suffices to express $w^{\mathrm{RL2ML}}_{\gamma,N}(p)$ in the Bernstein basis:
\begin{align*}
w^{\mathrm{RL2ML}}_{\gamma,N}(p)
=
\sum_{j=0}^{N-1}\beta_j^{(\gamma,N)}B_{j,N-1}(p).
\end{align*}
The corresponding coefficients admit the closed form
\begin{align*}
\beta_{K-1}^{(\gamma,N)}
=
\frac{\Gamma(N+\gamma)}{\Gamma(N)}
\frac{\Gamma(K)}{\Gamma(K+\gamma)},\ \text{for}\ K=1,\ldots,N.
\end{align*}
Equivalently, the group-level update scale is
\begin{align*}
\alpha_K^{(\gamma,N)}
=
\frac{K}{N}\beta_{K-1}^{(\gamma,N)}
=
\frac{\Gamma(N+\gamma)}{\Gamma(N+1)}
\frac{\Gamma(K+1)}{\Gamma(K+\gamma)}.
\end{align*}
The derivation is given in Appendix~\ref{app:derive-beta-closed-form}.

\begin{theorem}[Exactly unbiased estimator for the truncated RL2ML objective]
\label{thm:exact}
Define
\begin{align*}
\hat g_{\gamma,N}^{\mathrm{RL2ML}}(x)
=
\alpha_K^{(\gamma,N)}\bar S_K,\ \text{where}\ \alpha_K^{(\gamma,N)}
=
\frac{\Gamma(N+\gamma)}{\Gamma(N+1)}
\frac{\Gamma(K+1)}{\Gamma(K+\gamma)}.
\end{align*}
Then
\begin{align*}
\E[\hat g_{\gamma,N}^{\mathrm{RL2ML}}(x)\mid x]
=
\nabla_\theta\mathcal{J}^{\mathrm{RL2ML}}_{\gamma,N}(x).
\end{align*}
\end{theorem}

The proof is given in Appendix~\ref{app:proof-thm-exact}. The theorem shows that RL2ML is not merely a heuristic reweighting rule: for each fixed $\gamma$ and $N$, it defines a finite-rollout surrogate objective and an exactly unbiased estimator for that surrogate objective.

\subsubsection{Control Variate}
The estimator can be used directly as
\begin{align*}
\hat g_{\gamma,N}^{\mathrm{direct}}(x)
=
\frac1N\sum_{i=1}^N \beta_{K-1}^{(\gamma,N)}r_iS_i,\ \text{for}\ K\ge 1,
\end{align*}
with zero update when $K=0$. Since $\E[N^{-1}\sum_i S_i\mid x]=0$, subtracting $N^{-1}\sum_i S_i$ preserves unbiasedness while reducing variance:
\begin{align*}
\tilde g_{\gamma,N}(x)
=
\frac1N\sum_{i=1}^N\bigl(\beta_{K-1}^{(\gamma,N)}r_i-1\bigr)S_i,
\end{align*}
The sequence-level advantage for the control-variate form is
\begin{align*}
A_i^{(\gamma,N)}=\beta_{K-1}^{(\gamma,N)}r_i-1.
\end{align*}
For $\gamma=1$,
\begin{align*}
\beta_{K-1}^{(1,N)}=\frac{N}{K},\ \text{and}\ A_i^{(1,N)}=\frac{N}{K}r_i-1,
\end{align*}
which recovers the MaxRL control-variate advantage. The direct form and the control-variate form estimate the same population gradient in expectation, but they differ in finite-sample variance and in how all-failure groups are treated.

\subsection{Update-Scale Geometry Controlled by \texorpdfstring{$\gamma$}{gamma}}
The parameter $\gamma$ controls both the population-level weight and the sample-level update scale. From the closed form,
\begin{align*}
\frac{\alpha_{K+1}^{(\gamma,N)}}{\alpha_K^{(\gamma,N)}}
=
\frac{K+1}{K+\gamma}.
\end{align*}
Thus $\gamma=1$ is a structural boundary in the observed rollout-group geometry.

\begin{proposition}[Subcritical and supercritical update-scale regimes]
\label{prop:ratio}
For fixed $N$, the following statements hold:
\begin{enumerate}
    \item If $\gamma=0$, then $\alpha_K^{(0,N)}=K/N$, which is the ordinary RL.
    \item If $0<\gamma<1$, then $\alpha_K^{(\gamma,N)}$ is increasing in $K$ and satisfies $K/N<\alpha_K^{(\gamma,N)}<1$ for $1\le K<N$. This is the subcritical regime.
    \item If $\gamma=1$, then $\alpha_K^{(1,N)}\equiv 1$, which is the MaxRL boundary.
    \item If $\gamma>1$, then $\alpha_K^{(\gamma,N)}$ is decreasing in $K$ and satisfies $\alpha_K^{(\gamma,N)}>1$ for all $K<N$. This is the supercritical regime.
\end{enumerate}
\end{proposition}

\begin{figure}[htbp]
    \centering
    \includegraphics[width=16cm]{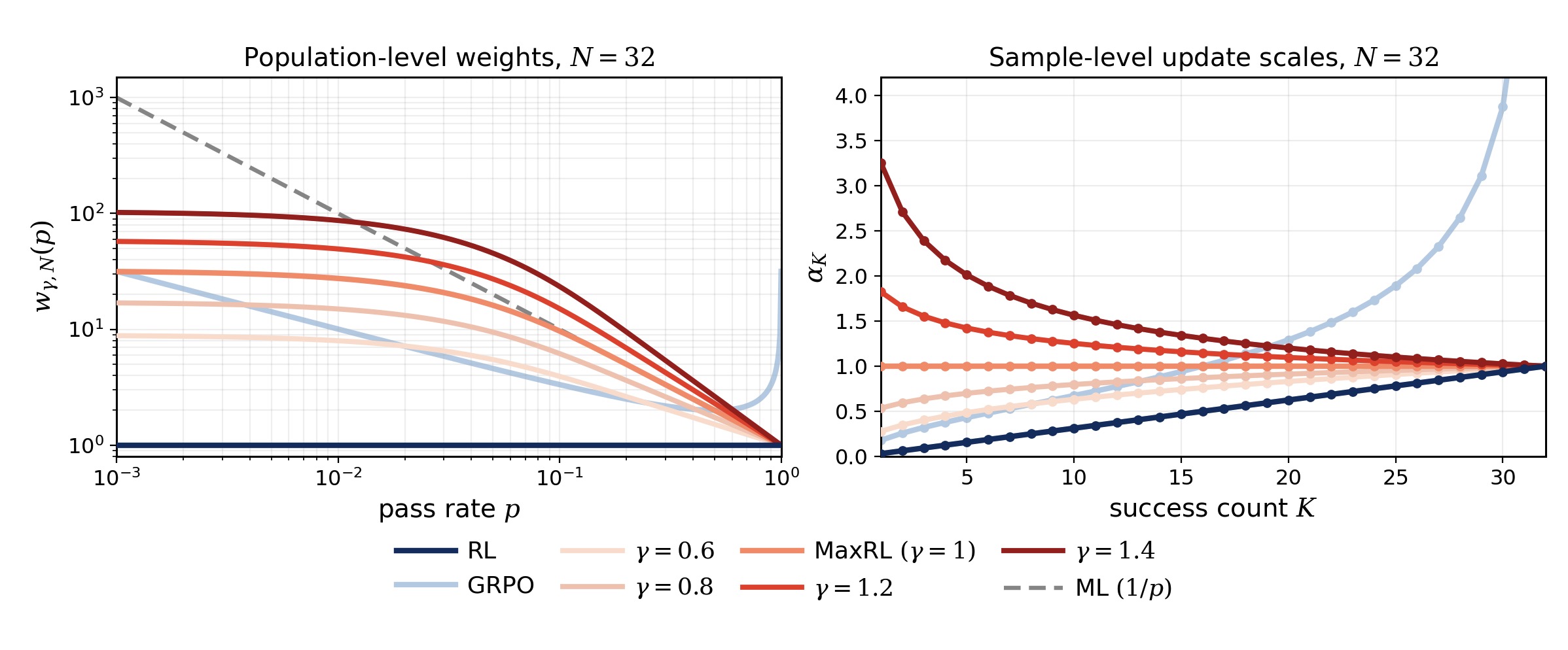}
    \caption[Population-level weights and sample-level update scales for RL2ML]{Population-level weights and sample-level update scales for RL2ML. The left panel shows the population-level weight $w_{\gamma,N}(p)$, while the right panel shows the sample-level update scale $\alpha_K$ for $N=32$. Values $\gamma<1$ form the subcritical regime, $\gamma=1$ is the MaxRL boundary, and $\gamma>1$ enters the supercritical low-$K$ amplification regime.}
    \label{fig:weights-alpha}
\end{figure}

The proof is given in Appendix~\ref{app:proof-prop-ratio}. The terms subcritical and supercritical refer to the update scale assigned to low-success rollout groups. Ordinary RL attenuates such groups in proportion to $K/N$; subcritical RL2ML still attenuates them but less severely; MaxRL removes the success-count dependence of the successful-score average; and supercritical RL2ML amplifies rare-success groups beyond the MaxRL scale. Figure~\ref{fig:weights-alpha} reports the population-level weight and the sample-level update scale in the two panels.

\section{Finite-Horizon Selection of \texorpdfstring{$\gamma$}{gamma}}
\label{sec:finite-horizon-selection}

The preceding sections define a family of rollout-budget-aligned surrogate objectives and exactly unbiased finite-rollout estimators. The remaining question is how to select the member of this family used for training. Throughout this section, $\gamma^*$ denotes the value of $\gamma$ chosen by an outer selection criterion over a candidate interval $\Gamma=[\gamma_{\min},\gamma_{\max}]$. A static rule such as choosing the surrogate closest to maximum likelihood is therefore insufficient. Finite-horizon RLVR is evaluated after a finite number of optimizer steps, and different choices of $gamma$ can induce different update magnitudes and different estimator noise. This section first constructs a calibrated metric-gain criterion, then incorporates a variance penalty, and finally defines $\gamma^*$ as the solution of the resulting one-dimensional outer optimization problem.

For notational simplicity, write
\begin{align*}
w_{\gamma,N}(p):=w_{\gamma,N}^{\mathrm{RL2ML}}(p)
=
\sum_{m=0}^{N-1}\frac{(\gamma)_m}{m!}(1-p)^m.
\end{align*}

\subsection{Why Effective Learning-Rate Calibration Is Necessary}
Changing $\gamma$ changes the coefficients $\beta_{K-1}^{(\gamma,N)}$, the group-level scale $\alpha_K^{(\gamma,N)}$, and the population-level weight $w_{\gamma,N}(p)$ simultaneously. If all values of $\gamma$ are compared under the same base learning rate, a larger observed improvement may simply reflect a larger effective step rather than a better surrogate objective, thereby conflating objective design with learning-rate scaling.

Accordingly, to compare surrogate objectives rather than update magnitudes, we normalize the expected update direction for each $\gamma$ and apply the same target update length $c$, so that all candidates are evaluated under the same effective parameter-update length.

Let $\mathcal C$ be the prompt set used for local analysis, and let the validation metric be prompt-separable:
\begin{align*}
V_{\mathcal C}(\theta)
=
\sum_{x\in\mathcal C}v_x(p_x),\ \text{where}\ p_x:=p_\theta(x).
\end{align*}
The function $v_x$ encodes the marginal value assigned by the actual evaluation metric to the success probability of prompt $x$. For example, pass@1 gives $v_x(p)=p$, pass@k gives $v_x(p)=1-(1-p)^k$, and a smoothed log-success metric gives $v_x(p)=\log(p+\tau)$.

For analytical clarity, we first consider a prompt-separable approximation in which the gradient contributions of different prompts are treated as orthogonal. Under this approximation, write $\theta=(\theta_x)_{x\in\mathcal C}$ and define the expected RL2ML update direction
\begin{align*}
D_{\gamma,\mathcal C}(\theta)
=
\sum_{x\in\mathcal C}w_{\gamma,N}(p_x)\nabla_{\theta_x}p_x,
\end{align*}
and define
\begin{align*}
\ell_x:=\|\nabla_{\theta_x}p_x\|_2^2.
\end{align*}
Here $\ell_x$ measures how strongly the success probability of prompt $x$ changes under its corresponding parameter block. Set
\begin{gather}
a_x:=v_x'(p_x)\ell_x,\ \text{and}\ b_x:=\ell_x,\notag\\
A(\gamma):=\sum_{x\in\mathcal C}a_xw_{\gamma,N}(p_x),\ \text{and}\ B(\gamma):=\sum_{x\in\mathcal C}b_xw_{\gamma,N}(p_x)^2.
\label{eq:ab-def}
\end{gather}
Here $A(\gamma)$ is the first-order alignment with the validation metric, while $B(\gamma)$ is the squared norm of the expected update direction. Given a target update length $c>0$, we define
\begin{align*}
\theta_\gamma^+
:=
\theta+\eta_\gamma D_{\gamma,\mathcal C}(\theta),\ \text{where}\ \eta_\gamma:=\frac{c}{\|D_{\gamma,\mathcal C}(\theta)\|_2}
=
\frac{c}{\sqrt{B(\gamma)}}.
\end{align*}
This gives $\|\theta_\gamma^+-\theta\|_2=c$, so different values of $\gamma$ are compared under the same parameter-update length.

The calibrated first-order metric-gain criterion is
\begin{align}
U(\gamma)
:=
\frac{A(\gamma)}{\sqrt{B(\gamma)}}
=
\frac{ \sum_{x\in\mathcal C} v_x'(p_x)\,w_{\gamma,N}(p_x)\,\ell_x }{ \sqrt{ \sum_{x\in\mathcal C} w_{\gamma,N}(p_x)^2\,\ell_x } }.
\label{eq:u-def}
\end{align}

\begin{proposition}[Calibrated metric gain under fixed update length]
\label{prop:localgain}
Under the prompt-separable approximation introduced above, the fixed-$c$ update satisfies
\begin{align*}
V_{\mathcal C}(\theta_\gamma^+)-V_{\mathcal C}(\theta)
=
c\,U(\gamma)+\mathcal{O}(c^2).
\end{align*}
\end{proposition}

The derivation is given in Appendix~\ref{app:proof-prop-localgain}. Here $\mathcal{O}(c^2)$ denotes a standard local Taylor remainder bounded by a constant times $c^2$ as the target update length size $c\to 0$.

This proposition gives the first reason why $\gamma$ cannot be selected only by proximity to maximum likelihood. Even if increasing $\gamma$ gives more population-level weight to low-$p$ prompts, the calibrated metric gain also depends jointly on the metric marginal value $v_x'(p_x)$, the population-level weight $w_{\gamma,N}(p_x)$, and the prompt sensitivity coefficient $\ell_x$. Therefore, prompts with large surrogate weight do not necessarily dominate the finite-horizon improvement: if they have weak prompt sensitivity or low marginal value under the evaluation metric, increasing $\gamma$ can reduce the gain per unit effective update length.

\begin{corollary}[A sufficient condition for improving over the MaxRL]\label{cor:supercritical}
Assume $A(1)>0$ and $B(1)>0$. If
\begin{align*}
\frac{B'(1)}{B(1)}
>
2\frac{A'(1)}{A(1)},
\end{align*}
then $U'(1)<0$. Hence there exists $\varepsilon>0$ such that $U(1-\delta)>U(1)$ for every $0<\delta<\varepsilon$.
\end{corollary}

\begin{figure}[htbp]
    \centering
    \includegraphics[width=14cm]{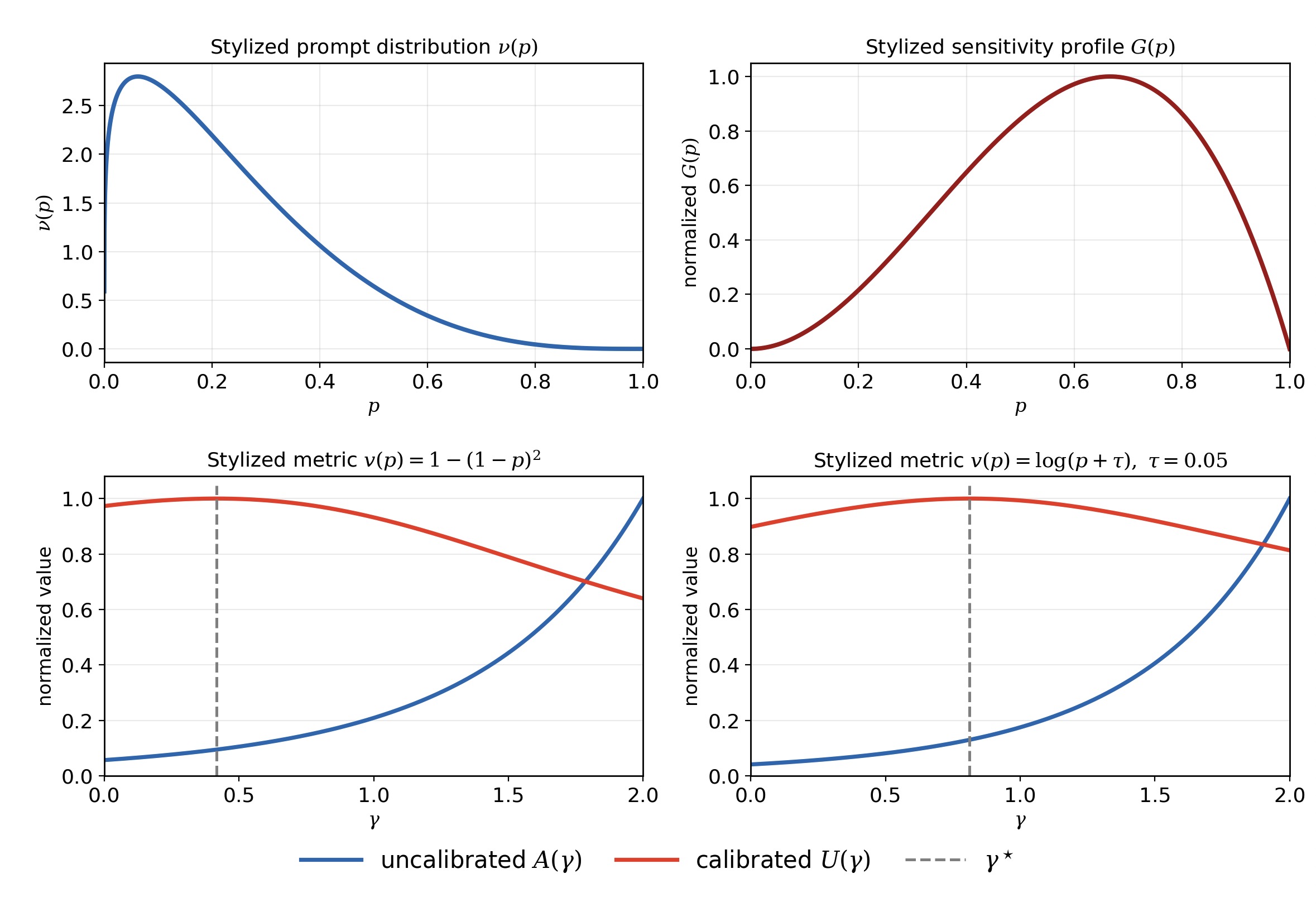}
    \caption[Stylized finite-horizon selection curves]{Stylized finite-horizon selection curves. This example illustrates that the best $\gamma$ is metric-dependent and calibration-dependent, rather than a static boundary point closest to $1/p$.}
    \label{fig:selector}
\end{figure}

The proof is given in Appendix~\ref{app:proof-prop-localgain}. This criterion formalizes when the MaxRL point can be locally improved by moving into the subcritical region under a given parameter state and evaluation metric. Figure~\ref{fig:selector} shows a stylized example in which the uncalibrated first-order metric gain $A(\gamma)$ favors larger $\gamma$, while the calibrated first-order metric-gain criterion $U(\gamma)$ can have an interior maximum.

\subsection{Estimator Variance and the Outer Objective}
Calibrated metric gain describes the expected first-order improvement. However, an update direction that has high expected gain can still be undesirable if its finite-rollout estimator is too noisy. 因此在选择$\gamma^*$时同样需要考虑estimator variance. Define
\begin{align*}
a_K(\gamma):=\alpha_K^{(\gamma,N)}\I\{K\ge 1\},\ \text{and}\ \hat g_{\gamma,N}(x)=a_K(\gamma)\bar S_K,
\end{align*}
with the convention that $a_K(\gamma)^2/K:=0$ when $K=0$.

\begin{theorem}[Conditional variance decomposition]
\label{thm:var}
For any fixed prompt $x$ and fixed $\gamma,N$,
\begin{align*}
\Cov(\hat g_{\gamma,N}(x)\mid x)
=
\Var(a_K(\gamma)\mid x)\mu_x\mu_x^\top
+
\E\!\left[\frac{a_K(\gamma)^2}{K}\middle|x\right]\Sigma_x.
\end{align*}
Consequently,
\begin{align*}
\E\!\left[\|\hat g_{\gamma,N}(x)-\E[\hat g_{\gamma,N}(x)\mid x]\|_2^2\middle|x\right]
=
\Var(a_K(\gamma)\mid x)\|\mu_x\|_2^2
+
\E\!\left[\frac{a_K(\gamma)^2}{K}\middle|x\right]\Tr(\Sigma_x).
\end{align*}
\end{theorem}

The proof is given in Appendix~\ref{app:proof-thm-var}. The first term is count variance caused by the random success count; the second is within-success variance caused by the randomness of successful trajectory score function. Both terms scale quadratically with $a_K(\gamma)$. In particular, the supercritical regime $\gamma>1$ carries an explicit finite-rollout noise cost because it assigns larger update scales to low-$K$ rollout groups.

Aggregating over the calibration prompt set $\mathcal C$ gives the variance proxy $R(\gamma)$:
\begin{align}
R(\gamma)
:=
\sum_{x\in\mathcal C}
\left[
\Var(a_K(\gamma)\mid x)\|\mu_x\|_2^2
+
\E\!\left[\frac{a_K(\gamma)^2}{K}\middle|x\right]\Tr(\Sigma_x)
\right].
\label{eq:r-def}
\end{align}
For theoretical evaluation, $K\mid x\sim\Binom(N,p_x)$ can be used in Eq.~\eqref{eq:r-def}. In practical controllers, $p_x$ and the norm factors are replaced by smoothed estimates and low-cost proxies.

\begin{figure}[htbp]
    \centering
    \includegraphics[width=14cm]{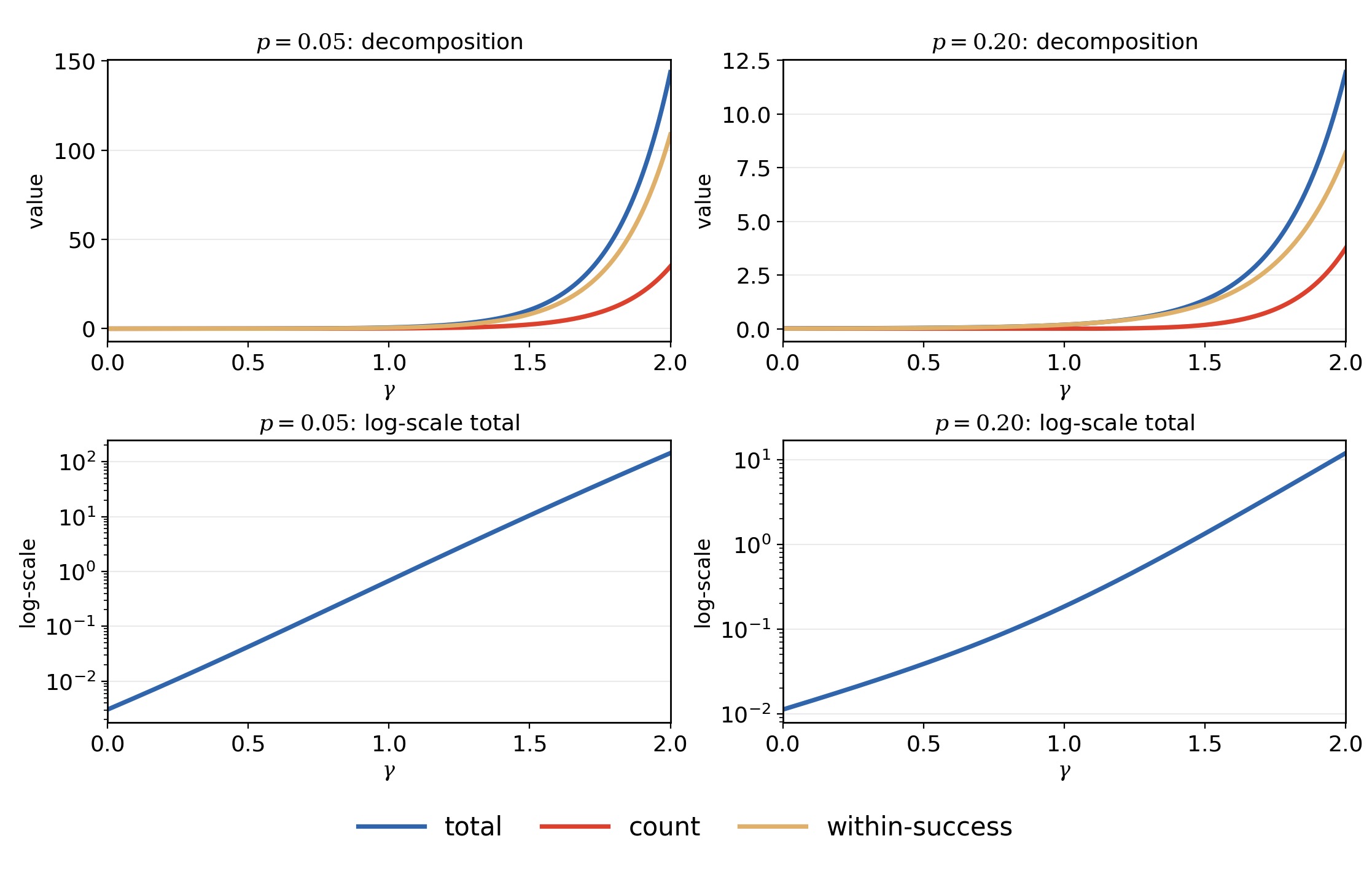}
    \caption[Noise curves induced by the exact variance decomposition]{Noise curves induced by the exact variance decomposition for $N=32$ under the normalization $\|\mu_x\|^2=\Tr(\Sigma_x)=1$. As $\gamma$ increases, both count variance and within-success variance increase, with faster growth at smaller success probabilities.}
    \label{fig:variance}
\end{figure}

Figure~\ref{fig:variance} visualizes the variance growth implied by Theorem~\ref{thm:var} in a normalized setting.

After effective learning-rate calibration, the variance-aware selection problem is
\begin{align}
\gamma_\lambda^*
=
\operatorname*{arg\,max}_{\gamma\in\Gamma}
\left[U(\gamma)-\lambda_{\mathrm{var}}R(\gamma)^{1/2}\right],\ \text{where}\ \Gamma=[\gamma_{\min},\gamma_{\max}]\ \text{and}\ \lambda_{\mathrm{var}}\ge 0.
\label{eq:gamma-star-main}
\end{align}
Equation~\eqref{eq:gamma-star-main} makes the outer choice explicit: $U(\gamma)$ measures expected metric improvement per unit effective update scale, while $R(\gamma)^{1/2}$ measures stochastic update noise. The coefficient $\lambda_{\mathrm{var}}$ reflects the tolerance for estimator variance. The derivative, first-order condition, and a Newton update for the unpenalized part $U(\gamma)$ are given once in Appendix~\ref{app:gamma-derivatives}.

This objective preserves the main conclusion of the paper. The selection of $\gamma$ is not equivalent to choosing the population weight closest to $1/p$, and it is not determined by static hard-prompt emphasis. It is a finite-horizon optimization problem governed by the evaluation metric $V$, the current prompt set $\mathcal C$, local sensitivity $\ell_x$, and estimator variance $R$.

\section{Related Work}
\label{sec:related-work}

\paragraph{Induced objectives for binary-reward reasoning RL.}
Davis and Recht provide a broader induced-objective view of binary-reward LLM reinforcement learning, showing that several algorithms can be interpreted as optimizing monotone transforms of prompt-level success probability\cite{objective}. RL2ML can be viewed as a specialization of this perspective: it studies the canonical power-likelihood path with population derivative $p^{-\gamma}$, connects this path to finite-rollout surrogate objectives, and derives a closed-form $K$-only unbiased estimator with a group-level update-scale transition at $\gamma=1$. Unlike Davis and Recht, RL2ML specifically organizes this family as a continuous path from RL to ML and beyond ML, and derives the associated coefficients, subcritical-supercritical transition, and variance-aware selection rule for $\gamma$. MaxRL is more directly adjacent because it derives a finite-compute surrogate for maximum-likelihood-style optimization in correctness-based RLVR\cite{maxrl}. RL2ML keeps the likelihood-style interpretation but does not assume that the objective closest to $\log p$ is always best under a finite training horizon. Instead, it separates the population-level weighting function from effective update scale and estimator variance, which makes maximum likelihood a reference point rather than a universal selection rule.

\paragraph{Policy-gradient estimators and group-based LLM RL.}
Classical policy-gradient methods estimate gradients by likelihood-ratio scores, with REINFORCE and the policy-gradient theorem providing the standard foundations\cite{williams1992reinforce,sutton1999policygradient}. PPO became a common optimizer for RLHF-style language-model post-training because it stabilizes policy updates through clipping and related trust-region heuristics\cite{schulman2017ppo,ouyang2022instructgpt}. More recent LLM post-training work revisits simpler REINFORCE-style estimators and group-based relative advantages, including RLOO-style methods and GRPO\cite{ahmadian2024rloo,shao2024deepseekmath,guo2025deepseekr1}. RL2ML is orthogonal to the choice of outer optimizer: it specifies the finite-rollout surrogate objective and the success-count-dependent scalar coefficient, whereas PPO, RLOO, and GRPO specify how sampled sequence advantages are converted into policy updates.

\section{Conclusion}
\label{sec:conclusion}

This paper introduced RL2ML as a finite-rollout surrogate-objective framework for correctness-based RLVR. Relative to prior induced-objective analyses and MaxRL, RL2ML preserves the estimator-objective alignment while exposing a continuous objective path from ordinary reinforcement learning, through the maximum-likelihood boundary, to supercritical objectives that reweight low-success prompts beyond maximum likelihood.

A central contribution of this work is the separation between population-level weighting and sample-level update geometry. The Bernstein representation characterizes what a $K$-only estimator optimizes in expectation, but the group-level update scale $\alpha_K$ determines how a realized rollout group is actually amplified or attenuated after its success count is observed. This distinction exposes the subcritical--supercritical transition: below the maximum-likelihood boundary, low-success groups remain attenuated relative to MaxRL; above it, they are amplified beyond the MaxRL scale. Thus, the effect of changing the surrogate objective is not fully captured by the static population-level weighting function alone.

The finite-horizon analysis further shows that proximity to maximum likelihood is not a sufficient principle for objective selection. After effective learning-rate calibration, the relevant comparison is the metric gain per unit update scale, not the uncalibrated size of the expected gradient. The calibrated metric-gain criterion makes the dependence on the validation metric and local sensitivity explicit, while the exact variance decomposition shows that supercritical low-success amplification carries a quantifiable finite-rollout noise cost. Consequently, neither ordinary reinforcement learning nor direct maximum-likelihood approximation is generally optimal across metrics and prompt distributions.

The resulting view treats the remaining degree of freedom in RL2ML as an outer-control variable rather than an unconstrained hyperparameter. In the oracle setting, it defines a one-dimensional variance-aware optimization problem balancing calibrated metric gain against estimator noise. In practical Verl-style training, the same structure can be approximated from recent on-policy success counts or from a calibration shard, then implemented by updating the group-level coefficient table together with effective learning-rate calibration. This provides a principled route for selecting a finite-rollout training surrogate for a given metric and parameter state, while keeping the underlying RLVR rollout and verification pipeline unchanged.

\clearpage
\appendix
\section*{Appendix}
\addcontentsline{toc}{section}{Appendix}

\section{Proofs and Derivations}
\label{app:proofs}

\subsection{Proof of Theorem~\ref{thm:bernstein}}
\label{app:proof-thm-bernstein}

\paragraph{Restatement of Theorem~\ref{thm:bernstein}.}
For a fixed prompt $x$, consider
\begin{align*}
\hat g_f(x)=f(K)\sum_{i=1}^N r_iS_i,\ \text{where}\ K=\sum_{i=1}^N r_i.
\end{align*}
With $\beta_m=Nf(m+1)$ and $B_{m,N-1}(p)=\binom{N-1}{m}p^m(1-p)^{N-1-m}$,
\begin{align*}
\E[\hat g_f(x)\mid x]
=
\left(\sum_{m=0}^{N-1}\beta_mB_{m,N-1}(p)\right)\nabla_\theta p.
\end{align*}

\paragraph{Proof.}
By exchangeability,
\begin{align*}
\E[\hat g_f(x)\mid x]
=
N\E[f(K)r_1S_1\mid x].
\end{align*}
Let $M_{-1}:=\sum_{j\neq 1}r_j$. Since $r_1S_1=0$ on $r_1=0$, the event contributing to the expectation has $K=1+M_{-1}$. The random variable $M_{-1}$ is conditionally independent of $r_1S_1$ given $x$, so
\begin{align*}
\E[f(K)r_1S_1\mid x]
=
\E[f(1+M_{-1})\mid x]\E[r_1S_1\mid x].
\end{align*}
Furthermore,
\begin{align*}
\E[r_1S_1\mid x]
&=
\sum_z m_\theta(z\mid x)r(x,z)\nabla_\theta\log m_\theta(z\mid x)\\
&=
\sum_z r(x,z)\nabla_\theta m_\theta(z\mid x)
=
\nabla_\theta p.
\end{align*}
Since $M_{-1}\sim\Binom(N-1,p)$,
\begin{align*}
\E[f(1+M_{-1})\mid x]
=
\sum_{m=0}^{N-1}f(m+1)\binom{N-1}{m}p^m(1-p)^{N-1-m}.
\end{align*}
Multiplying by $N\nabla_\theta p$ and using $\beta_m=Nf(m+1)$ gives the claimed Bernstein representation.

\subsection{Proof of Proposition~\ref{prop:gradexp}}
\label{app:proof-prop-gradexp}

\paragraph{Restatement of Proposition~\ref{prop:gradexp}.}
For
\begin{align*}
\mathcal{J}^{\mathrm{RL2ML}}_{\gamma,N}(x)
=
\sum_{k=1}^N\frac{(\gamma)_{k-1}}{(k-1)!\,k}\pass@k(x),
\end{align*}
we have
\begin{align*}
\nabla_\theta\mathcal{J}^{\mathrm{RL2ML}}_{\gamma,N}(x)
=
w^{\mathrm{RL2ML}}_{\gamma,N}(p_\theta(x))\nabla_\theta p_\theta(x).
\end{align*}

\paragraph{Proof.}
For fixed $x$, write $p=p_\theta(x)$. Since $\pass@k(x)=1-(1-p)^k$,
\begin{align*}
\nabla_\theta\pass@k(x)
=
k(1-p)^{k-1}\nabla_\theta p.
\end{align*}
Substitution gives
\begin{align*}
\nabla_\theta\mathcal{J}^{\mathrm{RL2ML}}_{\gamma,N}(x)
&=
\sum_{k=1}^N\frac{(\gamma)_{k-1}}{(k-1)!\,k}k(1-p)^{k-1}\nabla_\theta p\\
&=
\left(\sum_{m=0}^{N-1}\frac{(\gamma)_m}{m!}(1-p)^m\right)\nabla_\theta p.
\end{align*}
The term in parentheses is $w^{\mathrm{RL2ML}}_{\gamma,N}(p)$.

\subsection{Bernstein Expansion and Closed-Form Coefficients}
\label{app:derive-beta-closed-form}
This subsection combines the Bernstein-basis conversion and the coefficient simplification. For $m=0,\ldots,N-1$,
\begin{align*}
(1-p)^m
=
\sum_{j=0}^{N-1-m}
\frac{\binom{N-1-j}{m}}{\binom{N-1}{m}}B_{j,N-1}(p).
\end{align*}
To verify the identity, expand the right-hand side and use
\begin{align*}
\binom{N-1-j}{m}\binom{N-1}{j}
=
\binom{N-1}{m}\binom{N-1-m}{j}.
\end{align*}
The remaining sum is the binomial expansion of $(p+(1-p))^{N-1-m}$ after factoring out $(1-p)^m$.

Using this identity in
\begin{align*}
w^{\mathrm{RL2ML}}_{\gamma,N}(p)
=
\sum_{m=0}^{N-1}\frac{(\gamma)_m}{m!}(1-p)^m
\end{align*}
and grouping terms by $B_{j,N-1}(p)$ yields
\begin{align*}
\beta_j^{(\gamma,N)}
=
\sum_{m=0}^{N-1-j}
\frac{(\gamma)_m}{m!}
\frac{\binom{N-1-j}{m}}{\binom{N-1}{m}}.
\end{align*}
Set $j=K-1$. Then
\begin{align*}
\beta_{K-1}^{(\gamma,N)}
=
\sum_{m=0}^{N-K}
\frac{(\gamma)_m}{m!}
\frac{\binom{N-K}{m}}{\binom{N-1}{m}}.
\end{align*}
Using $\binom{A}{m}=(-A)_m/((-1)^m m!)$ for nonnegative integer $A$ gives
\begin{align*}
\beta_{K-1}^{(\gamma,N)}
=
{}_2F_1\bigl(-(N-K),\gamma;-(N-1);1\bigr).
\end{align*}
By the Chu--Vandermonde identity\cite{bailey1935,koepf1998},
\begin{align*}
{}_2F_1(-n,b;c;1)=\frac{(c-b)_n}{(c)_n},
\end{align*}
so with $n=N-K$, $b=\gamma$, and $c=-(N-1)$,
\begin{align*}
\beta_{K-1}^{(\gamma,N)}
=
\frac{(-(N-1)-\gamma)_{N-K}}{(-(N-1))_{N-K}}.
\end{align*}
The identity $(-a)_n=(-1)^n\Gamma(a+1)/\Gamma(a-n+1)$ gives
\begin{align*}
\beta_{K-1}^{(\gamma,N)}
=
\frac{\Gamma(N+\gamma)}{\Gamma(N)}
\frac{\Gamma(K)}{\Gamma(K+\gamma)}.
\end{align*}
At $\gamma=1$, this reduces to $\beta_{K-1}^{(1,N)}=N/K$, recovering the MaxRL coefficient.

\subsection{Proof of Theorem~\ref{thm:exact}}
\label{app:proof-thm-exact}

\paragraph{Restatement of Theorem~\ref{thm:exact}.}
Define
\begin{align*}
\hat g_{\gamma,N}^{\mathrm{RL2ML}}(x)
=
\alpha_K^{(\gamma,N)}\bar S_K,\ \text{where}\ \alpha_K^{(\gamma,N)}
=
\frac{\Gamma(N+\gamma)}{\Gamma(N+1)}
\frac{\Gamma(K+1)}{\Gamma(K+\gamma)}.
\end{align*}
Then
\begin{align*}
\E[\hat g_{\gamma,N}^{\mathrm{RL2ML}}(x)\mid x]
=
\nabla_\theta\mathcal{J}^{\mathrm{RL2ML}}_{\gamma,N}(x).
\end{align*}

\paragraph{Proof.}
For $K\ge 1$,
\begin{align*}
\hat g_{\gamma,N}^{\mathrm{RL2ML}}(x)
=
\alpha_K^{(\gamma,N)}\frac1K\sum_{i=1}^N r_iS_i
=
\frac{\beta_{K-1}^{(\gamma,N)}}{N}\sum_{i=1}^N r_iS_i.
\end{align*}
This is a $K$-only estimator of the form in Theorem~\ref{thm:bernstein}. Therefore,
\begin{align*}
\E[\hat g_{\gamma,N}^{\mathrm{RL2ML}}(x)\mid x]
=
\left(\sum_{j=0}^{N-1}\beta_j^{(\gamma,N)}B_{j,N-1}(p)\right)\nabla_\theta p.
\end{align*}
By Appendix~\ref{app:derive-beta-closed-form}, the coefficients are exactly those of $w_{\gamma,N}^{\mathrm{RL2ML}}(p)$, and Proposition~\ref{prop:gradexp} gives the desired identity.

\subsection{Proof of Proposition~\ref{prop:ratio}}
\label{app:proof-prop-ratio}

\paragraph{Restatement of Proposition~\ref{prop:ratio}.}
For fixed $N$, $\alpha_K^{(0,N)}=K/N$ at $\gamma=0$; $\alpha_K^{(\gamma,N)}$ is increasing in $K$ and satisfies $K/N<\alpha_K^{(\gamma,N)}<1$ for $0<\gamma<1$ and $1\le K<N$; $\alpha_K^{(1,N)}\equiv 1$ at the MaxRL boundary; and $\alpha_K^{(\gamma,N)}$ is decreasing in $K$ and satisfies $\alpha_K^{(\gamma,N)}>1$ for $\gamma>1$ and $K<N$.

\paragraph{Proof.}
From the closed form of $\alpha_K^{(\gamma,N)}$,
\begin{align*}
\frac{\alpha_{K+1}^{(\gamma,N)}}{\alpha_K^{(\gamma,N)}}
=
\frac{\Gamma(K+2)\Gamma(K+\gamma)}{\Gamma(K+1)\Gamma(K+1+\gamma)}
=
\frac{K+1}{K+\gamma}.
\end{align*}
If $0\le\gamma<1$, this ratio is larger than one and $\alpha_N^{(\gamma,N)}=1$, so $\alpha_K^{(\gamma,N)}$ is increasing in $K$ and does not exceed one. The case $\gamma=0$ gives $\alpha_K^{(0,N)}=K/N$. If $\gamma=1$, the ratio is one and $\alpha_K^{(1,N)}\equiv 1$. If $\gamma>1$, the ratio is smaller than one and $\alpha_N^{(\gamma,N)}=1$, so $\alpha_K^{(\gamma,N)}>1$ for $K<N$.

\subsection{Proof of Proposition~\ref{prop:localgain} and Corollary~\ref{cor:supercritical}}
\label{app:proof-prop-localgain}

\paragraph{Restatement of Proposition~\ref{prop:localgain}.}
Under the prompt-separable local model, the matched-norm update $\theta_\gamma^+=\theta+\eta_\gamma D_{\gamma,\mathcal C}(\theta)$ with $\eta_\gamma=c/\|D_{\gamma,\mathcal C}(\theta)\|_2$ satisfies
\begin{align*}
V_{\mathcal C}(\theta_\gamma^+)-V_{\mathcal C}(\theta)
=
c\,U(\gamma)+\mathcal{O}(c^2).
\end{align*}

\paragraph{Restatement of Corollary~\ref{cor:supercritical}.}
If $A(1)>0$, $B(1)>0$, and $B'(1)/B(1)>2A'(1)/A(1)$, then $U'(1)<0$ and there exists $\varepsilon>0$ such that $U(1-\delta)>U(1)$ for every $0<\delta<\varepsilon$.

\paragraph{Proof.}
Under the prompt-separable local model,
\begin{align*}
\nabla_{\theta_x}V_{\mathcal C}(\theta)
=
v_x'(p_x)\nabla_{\theta_x}p_x.
\end{align*}
Orthogonality of different prompt blocks gives
\begin{align*}
\langle\nabla V_{\mathcal C}(\theta),D_{\gamma,\mathcal C}(\theta)\rangle
=
\sum_{x\in\mathcal C}v_x'(p_x)w_{\gamma,N}(p_x)\|\nabla_{\theta_x}p_x\|_2^2
=
A(\gamma),
\end{align*}
and
\begin{align*}
\|D_{\gamma,\mathcal C}(\theta)\|_2^2
=
\sum_{x\in\mathcal C}w_{\gamma,N}(p_x)^2\|\nabla_{\theta_x}p_x\|_2^2
=
B(\gamma).
\end{align*}
A first-order Taylor expansion at $\theta$ gives
\begin{align*}
V_{\mathcal C}(\theta+\eta_\gamma D_{\gamma,\mathcal C})-V_{\mathcal C}(\theta)
=
\eta_\gamma A(\gamma)+\mathcal{O}(\eta_\gamma^2B(\gamma)).
\end{align*}
With $\eta_\gamma=c/\sqrt{B(\gamma)}$, the leading term is $cA(\gamma)/\sqrt{B(\gamma)}=cU(\gamma)$ and the remainder is $\mathcal{O}(c^2)$.

For the corollary, differentiate $U(\gamma)=A(\gamma)B(\gamma)^{-1/2}$:
\begin{align*}
U'(\gamma)
=
\frac{2A'(\gamma)B(\gamma)-A(\gamma)B'(\gamma)}{2B(\gamma)^{3/2}}.
\end{align*}
Under the stated condition at $\gamma=1$, the numerator is negative, so $U'(1)<0$. Continuity implies $U(1-\delta)>U(1)$ for all sufficiently small $\delta>0$.

\subsection{Derivatives of the Calibrated Objective}
\label{app:gamma-derivatives}
For $\gamma>0$,
\begin{align*}
\partial_\gamma(\gamma)_m
=
(\gamma)_m\bigl(\psi(\gamma+m)-\psi(\gamma)\bigr),
\end{align*}
where $\psi$ is the digamma function. Hence
\begin{align*}
\dot w_{\gamma,N}(p)
:=
\partial_\gamma w_{\gamma,N}(p)
=
\sum_{m=1}^{N-1}
\frac{(\gamma)_m}{m!}
\bigl(\psi(\gamma+m)-\psi(\gamma)\bigr)(1-p)^m.
\end{align*}
Let $\psi_1$ be the trigamma function. Then
\begin{align*}
\ddot w_{\gamma,N}(p)
:=
\partial_{\gamma\gamma}w_{\gamma,N}(p)
=
\sum_{m=1}^{N-1}
\frac{(\gamma)_m}{m!}
\left[
\bigl(\psi(\gamma+m)-\psi(\gamma)\bigr)^2
+
\psi_1(\gamma+m)-\psi_1(\gamma)
\right](1-p)^m.
\end{align*}
From Eq.~\eqref{eq:ab-def},
\begin{align*}
A'(\gamma)&=\sum_{x\in\mathcal C}a_x\dot w_{\gamma,N}(p_x),\\
A''(\gamma)&=\sum_{x\in\mathcal C}a_x\ddot w_{\gamma,N}(p_x),\\
B'(\gamma)&=2\sum_{x\in\mathcal C}b_xw_{\gamma,N}(p_x)\dot w_{\gamma,N}(p_x),\\
B''(\gamma)&=2\sum_{x\in\mathcal C}b_x\left[\dot w_{\gamma,N}(p_x)^2+w_{\gamma,N}(p_x)\ddot w_{\gamma,N}(p_x)\right].
\end{align*}
By Eq.~\eqref{eq:u-def},
\begin{align*}
U'(\gamma)
=
\frac{2A'(\gamma)B(\gamma)-A(\gamma)B'(\gamma)}{2B(\gamma)^{3/2}}.
\end{align*}
Stationary points of $U$ satisfy $F(\gamma)=0$, where
\begin{align*}
F(\gamma):=2A'(\gamma)B(\gamma)-A(\gamma)B'(\gamma).
\end{align*}
A projected Newton step is
\begin{align*}
\gamma
\leftarrow
\Pi_{[\gamma_{\min},\gamma_{\max}]}
\left(\gamma-\frac{F(\gamma)}{F'(\gamma)}\right).
\end{align*}
with
\begin{align*}
F'(\gamma)
=
2A''(\gamma)B(\gamma)+A'(\gamma)B'(\gamma)-A(\gamma)B''(\gamma).
\end{align*}
If $\gamma=0$ is included in the candidate interval, it is usually evaluated as a boundary point, while Newton steps are run with $\gamma_{\min}>0$. Since $U$ need not be globally concave, practical selection should compare boundary points, a grid of candidates, and stationary points found from multiple initializations.

\subsection{Proof of Theorem~\ref{thm:var}}
\label{app:proof-thm-var}

\paragraph{Restatement of Theorem~\ref{thm:var}.}
For $a_K(\gamma)=\alpha_K^{(\gamma,N)}\I\{K\ge 1\}$ and $\hat g_{\gamma,N}(x)=a_K(\gamma)\bar S_K$,
\begin{align*}
\Cov(\hat g_{\gamma,N}(x)\mid x)
=
\Var(a_K(\gamma)\mid x)\mu_x\mu_x^\top
+
\E\!\left[\frac{a_K(\gamma)^2}{K}\middle|x\right]\Sigma_x.
\end{align*}

\paragraph{Proof.}
Conditional on $K=k\ge 1$, the $k$ successful samples are conditionally i.i.d. from the success-conditioned rollout distribution. Therefore,
\begin{align*}
\E[\bar S_K\mid x,K=k]=\mu_x,\ \text{and}\ \Cov(\bar S_K\mid x,K=k)=\frac{\Sigma_x}{k}.
\end{align*}
Thus
\begin{align*}
\E[\hat g_{\gamma,N}\mid x,K=k]&=a_k(\gamma)\mu_x,\\
\Cov(\hat g_{\gamma,N}\mid x,K=k)&=\frac{a_k(\gamma)^2}{k}\Sigma_x.
\end{align*}
The total covariance identity gives
\begin{align*}
\Cov(\hat g_{\gamma,N}\mid x)
=
\Cov(\E[\hat g_{\gamma,N}\mid x,K]\mid x)
+
\E[\Cov(\hat g_{\gamma,N}\mid x,K)\mid x].
\end{align*}
The first term is $\Var(a_K(\gamma)\mid x)\mu_x\mu_x^\top$, and the second is $\E[a_K(\gamma)^2/K\mid x]\Sigma_x$. Taking the trace gives the squared-norm expression in Theorem~\ref{thm:var}.

\section{Fixed-Gamma Estimation and Practical \texorpdfstring{$\gamma^*$}{gamma*} Selection in Verl}
\label{app:verl}
This appendix consolidates the implementation details that are not needed in the main theoretical development. The purpose is to specify where the fixed-$\gamma$ RL2ML estimator enters a Verl-style on-policy RLVR loop and how the practical $\gamma^*$ controller can collect success counts from recent rollouts or from a calibration shard.

\subsection{Fixed-\texorpdfstring{$\gamma$}{gamma} Advantage Construction}
In Verl-style on-policy RLVR training\cite{verl2025,yu2025dapo}, RL2ML is implemented at the group-level advantage-estimator layer. The reward function still returns binary correctness, and the rollout pipeline is unchanged. For a fixed $\gamma$, the only additional object is the coefficient table $\{\beta_{K-1}^{(\gamma,N)}\}_{K=1}^N$ indexed by the prompt-level success count.

\begin{algorithm}[H]
\caption{Fixed-$\gamma$ RL2ML advantage construction in Verl}
\begin{algorithmic}[1]
\Require Current model parameters $\theta_t$, rollout budget $N$, fixed $\gamma$, coefficient table $\{\beta_{K-1}^{(\gamma,N)}\}_{K=1}^N$
\For{each prompt group $x$ in the current on-policy batch}
  \State Sample $N$ rollouts $z_1,\ldots,z_N\sim m_{\theta_t}(\cdot\mid x)$
  \State Compute binary rewards $r_i=\I\{f(z_i)=y^*(x)\}$
  \State Compute $K_x=\sum_{i=1}^N r_i$
  \If{using the direct estimator}
      \State Set $A_i=0$ for all $i$ if $K_x=0$; otherwise set $A_i=\beta_{K_x-1}^{(\gamma,N)}r_i$
  \Else
      \State Use the score-baseline control variate and set $A_i=\beta_{K_x-1}^{(\gamma,N)}r_i-1$, with $\beta_{K_x-1}^{(\gamma,N)}r_i=0$ when $K_x=0$
  \EndIf
  \State Broadcast $A_i$ to response tokens using the response mask
\EndFor
\State Apply the standard on-policy policy loss with the constructed token-level advantages
\end{algorithmic}
\end{algorithm}

\begin{lstlisting}[caption={RL2ML coefficient table and sequence-level advantages}]
import numpy as np
from scipy.special import gammaln


def precompute_beta_table(gamma: float, N: int) -> np.ndarray:
    """Return beta[K - 1] for K = 1,...,N."""
    K = np.arange(1, N + 1, dtype=np.float64)
    log_beta = (
        gammaln(N + gamma) - gammaln(N)
        + gammaln(K) - gammaln(K + gamma)
    )
    return np.exp(log_beta)


def compute_group_advantages(group_rewards: np.ndarray,
                             beta_table: np.ndarray,
                             use_control_variate: bool = True) -> np.ndarray:
    """group_rewards has shape [num_prompts, N] and values in {0,1}."""
    num_prompts, N = group_rewards.shape
    seq_adv = np.zeros_like(group_rewards, dtype=np.float32)
    for b in range(num_prompts):
        r = group_rewards[b].astype(np.float64)
        K = int(r.sum())
        if use_control_variate:
            if K == 0:
                seq_adv[b, :] = -1.0
            else:
                seq_adv[b, :] = beta_table[K - 1] * r - 1.0
        else:
            if K == 0:
                seq_adv[b, :] = 0.0
            else:
                seq_adv[b, :] = beta_table[K - 1] * r
    return seq_adv


def broadcast_to_tokens(seq_adv: np.ndarray,
                        response_mask: np.ndarray) -> np.ndarray:
    """seq_adv: [B, N], response_mask: [B, N, T]."""
    return seq_adv[:, :, None] * response_mask
\end{lstlisting}

\subsection{Collecting \texorpdfstring{$\{K_x\}$}{Kx} from an On-Policy Batch or Calibration Shard}
The controller requires prompt-level success counts under the current policy. For prompt $x$, this statistic is
\begin{align*}
K_x:=\sum_{i=1}^N r_{x,i},
\end{align*}
where the $N$ responses are the rollout group generated for the same prompt. In a recent on-policy batch, $K_x$ should be computed after the verifier has produced binary sequence-level rewards and before response-level examples are flattened irreversibly. If responses for the same prompt are contiguous, the reward vector can be reshaped to $[-1,N]$. If the training recipe shuffles responses, a prompt identifier should be retained and used for grouping.

A calibration shard uses the same statistic but collects it without optimizer steps. The current policy is rolled out on a fixed set of prompts, the verifier assigns binary rewards, and the resulting success counts are used only to evaluate the one-dimensional outer objective. This avoids comparing different training trajectories and keeps the controller tied to the current policy state.

\begin{lstlisting}[caption={Collecting prompt-level success counts from Verl-style tensors}]
import numpy as np
from collections import defaultdict


def collect_K_contiguous(sequence_rewards: np.ndarray, N: int) -> np.ndarray:
    """Use when each prompt has exactly N contiguous responses."""
    rewards = np.asarray(sequence_rewards, dtype=np.float64)
    assert rewards.ndim == 1
    assert rewards.size % N == 0
    group_rewards = rewards.reshape(-1, N)
    return group_rewards.sum(axis=1).astype(np.int64)


def collect_K_by_prompt_id(sequence_rewards: np.ndarray,
                           prompt_ids: np.ndarray,
                           N: int) -> np.ndarray:
    """Use when responses may not be contiguous but prompt ids are retained."""
    bucket = defaultdict(list)
    for pid, reward in zip(prompt_ids, sequence_rewards):
        bucket[int(pid)].append(float(reward))

    K_values = []
    for pid, rewards in bucket.items():
        if len(rewards) != N:
            raise ValueError(f"prompt {pid} has {len(rewards)} responses, expected {N}")
        K_values.append(int(np.sum(rewards)))
    return np.asarray(K_values, dtype=np.int64)


def collect_K_from_calibration_shard(policy, prompts, reward_fn, N: int) -> np.ndarray:
    """Sketch: run the current policy on a fixed shard without optimizer steps."""
    K_values = []
    for prompt in prompts:
        responses = policy.generate(prompt, n=N)
        rewards = np.asarray([reward_fn(prompt, y) for y in responses], dtype=np.float64)
        K_values.append(int(rewards.sum()))
    return np.asarray(K_values, dtype=np.int64)
\end{lstlisting}

\subsection{Practical Determination of \texorpdfstring{$\gamma^*$}{gamma*} in Verl}
Given success counts $K_x$, use a smoothed success-probability estimate
\begin{align*}
\hat p_x:=\frac{K_x+a}{N+a+b},
\end{align*}
with small Beta-prior parameters such as $a=b=1$. The metric marginal $v_x'(\hat p_x)$ is chosen from the validation metric: pass@1 gives $1$, pass@k gives $k(1-\hat p_x)^{k-1}$, and a smoothed log-success metric gives $(\hat p_x+\tau)^{-1}$. The local sensitivity scalar $\ell_x$ can be approximated by $\hat p_x(1-\hat p_x)$, by an exponential moving average of per-prompt policy-gradient norms, or by a task-specific reliability weight. The first choice is crude but stable and requires only $K_x$.

The practical controller evaluates $U(\gamma)$ and optionally the plug-in variance proxy $R(\gamma)$ on a one-dimensional grid, refines the best candidates using a safeguarded line search, and applies the selected $\gamma^*$ to construct the RL2ML coefficient table for the next training window. The effective learning rate should be calibrated using the token-level advantage RMS
\begin{align*}
\ARMS_\gamma
:=
\sqrt{
\frac{1}{\sum_{i=1}^{\mathcal R}T_i}
\sum_{i=1}^{\mathcal R}\sum_{t=1}^{T_i}
\bigl(A_{i,t}^{(\gamma)}\bigr)^2
},
\end{align*}
where $\mathcal R$ is the set of response sequences used for RMS estimation, $T_i$ counts response tokens only, and $\varepsilon_{\mathrm{rms}}>0$ is a numerical stabilizer. Relative to a reference $\gamma_{\mathrm{ref}}$,
\begin{align*}
\eta^{\mathrm{base}}_\gamma
=
\eta^{\mathrm{base}}_{\mathrm{ref}}
\frac{\mathrm{EMA}[\ARMS_{\gamma_{\mathrm{ref}}}]}
{\mathrm{EMA}[\ARMS_\gamma]+\varepsilon_{\mathrm{rms}}}.
\end{align*}
For offline comparison, the EMA can be replaced by the average on a calibration shard. For online adaptation, updating $\gamma$ less frequently than the policy parameters and clipping the learning-rate multiplier reduce sensitivity to count noise.

\begin{lstlisting}[caption={Online gamma selection from recent success counts}]
import numpy as np
from scipy.special import psi, polygamma


def weight_and_derivs(gamma: float, N: int, p: np.ndarray):
    one_minus_p = 1.0 - p
    w = np.ones_like(p, dtype=np.float64)
    dw = np.zeros_like(p, dtype=np.float64)
    ddw = np.zeros_like(p, dtype=np.float64)

    coeff = 1.0
    for m in range(1, N):
        coeff *= (gamma + m - 1.0) / m
        delta_psi = psi(gamma + m) - psi(gamma)
        delta_tri = polygamma(1, gamma + m) - polygamma(1, gamma)
        basis = one_minus_p ** m
        w += coeff * basis
        dw += coeff * delta_psi * basis
        ddw += coeff * (delta_psi**2 + delta_tri) * basis
    return w, dw, ddw


def gamma_objective(gamma, N, p_hat, ell_hat, vprime, eps=1e-8):
    w, dw, ddw = weight_and_derivs(gamma, N, p_hat)
    A = np.sum(vprime * w * ell_hat)
    B = np.sum((w**2) * ell_hat) + eps
    Ap = np.sum(vprime * dw * ell_hat)
    Bp = 2.0 * np.sum(w * dw * ell_hat)
    App = np.sum(vprime * ddw * ell_hat)
    Bpp = 2.0 * np.sum((dw**2 + w * ddw) * ell_hat)

    U = A / np.sqrt(B)
    F = 2.0 * Ap * B - A * Bp
    Fp = 2.0 * App * B + Ap * Bp - A * Bpp
    return U, F, Fp


def metric_marginal(p_hat: np.ndarray,
                    mode: str = "pass1",
                    pass_k: int = 1,
                    tau: float = 0.05) -> np.ndarray:
    if mode == "pass1":
        return np.ones_like(p_hat)
    if mode == "passk":
        return pass_k * (1.0 - p_hat) ** (pass_k - 1)
    if mode == "logp":
        return 1.0 / (p_hat + tau)
    raise ValueError(f"unknown metric mode: {mode}")


def select_gamma_from_counts(K_counts: np.ndarray,
                             N: int,
                             gamma_init: float = 0.8,
                             gamma_min: float = 1e-3,
                             gamma_max: float = 1.5,
                             a: float = 1.0,
                             b: float = 1.0,
                             metric_mode: str = "pass1",
                             pass_k: int = 1,
                             iters: int = 8) -> float:
    p_hat = (K_counts.astype(np.float64) + a) / (N + a + b)
    ell_hat = p_hat * (1.0 - p_hat)
    vprime = metric_marginal(p_hat, mode=metric_mode, pass_k=pass_k)

    grid = np.linspace(gamma_min, gamma_max, 41)
    values = [gamma_objective(g, N, p_hat, ell_hat, vprime)[0] for g in grid]
    gamma = float(grid[int(np.argmax(values))])
    gamma = float(np.clip(0.5 * gamma + 0.5 * gamma_init, gamma_min, gamma_max))

    for _ in range(iters):
        _, F, Fp = gamma_objective(gamma, N, p_hat, ell_hat, vprime)
        if abs(Fp) < 1e-10:
            break
        step = F / Fp
        candidate = float(np.clip(gamma - step, gamma_min, gamma_max))
        if not np.isfinite(candidate):
            break
        gamma = candidate
    return gamma
\end{lstlisting}
\section{Triad Family, the Role of \texorpdfstring{$M$}{M}, and Frontier Theory}
\label{app:triad}
The main text fixes $M=N$ to keep RL2ML a one-parameter finite-rollout family. A more general triad family introduces a truncation order $M\le N$ in addition to $\gamma$ and $N$. This extension is useful for understanding fidelity--stability tradeoffs, but it is not needed for the main estimator or for the one-dimensional $\gamma$ selection rule.

\subsection{General Triad Family}
Define
\begin{align*}
\mathcal{J}_{\gamma,M}(x)
=
\sum_{k=1}^M\frac{(\gamma)_{k-1}}{(k-1)!\,k}\pass@k(x),\ \text{for}\ M\le N.
\end{align*}
The corresponding exactly unbiased estimator has the form
\begin{align*}
\hat g_{\gamma,M,N}(x)
=
\alpha_K^{(\gamma,M,N)}\bar S_K,
\end{align*}
where
\begin{align*}
\alpha_K^{(\gamma,M,N)}
=
\frac{K}{N}
\sum_{m=0}^{\min(M-1,N-K)}
\frac{(\gamma)_m}{m!}
\frac{\binom{N-K}{m}}{\binom{N-1}{m}}.
\end{align*}
Thus $M$ changes the tail depth of the finite-compute surrogate objective, not the basic estimator class.

\subsection{Why \texorpdfstring{$M$}{M} Moves to the Boundary Without a Penalty}
For fixed $\gamma>0$,
\begin{align*}
w_{\gamma,M+1}(p)-w_{\gamma,M}(p)
=
\frac{(\gamma)_M}{M!}(1-p)^M>0.
\end{align*}
Therefore, if the outer criterion rewards larger static weights without any stability or optimization cost, the optimal $M$ is pushed to the boundary $M=N$. Values $M<N$ become meaningful only when fidelity is traded against stability, tail amplification, or finite-horizon optimization constraints.

\subsection{Frontier Theory}
\begin{figure}[htbp]
    \centering
    \includegraphics[width=12cm]{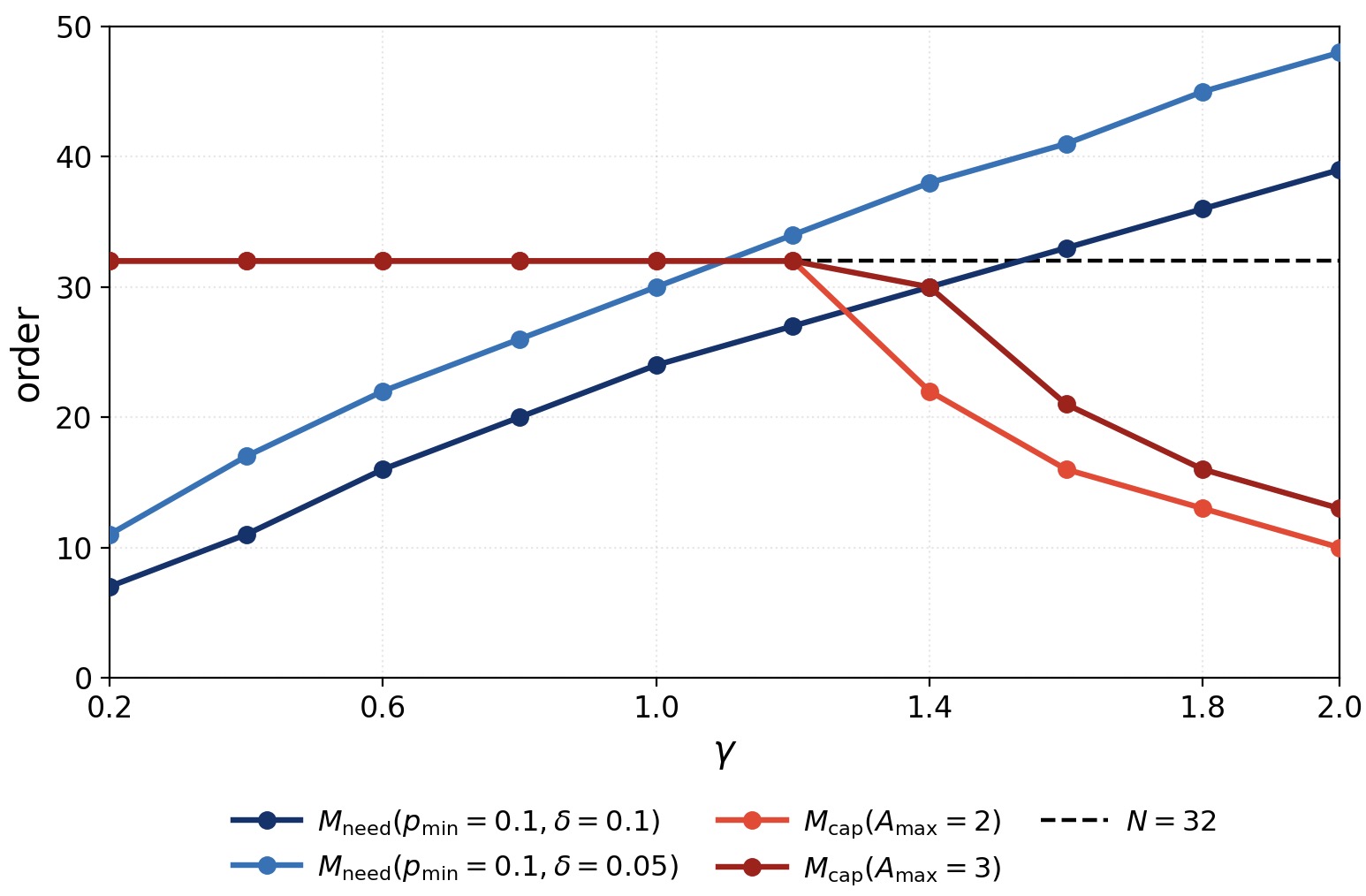}
    \caption[A representative triad-family frontier]{A representative triad-family frontier for $N=32$. The fidelity requirement determines $M_{\mathrm{need}}$, while the low-$K$ stability requirement determines $M_{\mathrm{cap}}$. The truncation order $M$ is useful as an independent degree of freedom only when the feasible interval is nonempty.}
    \label{fig:frontier}
\end{figure}

The infinite-order target is $p^{-\gamma}$, and the truncation tail is
\begin{align*}
R_{\gamma,M}(p)
:=
p^{-\gamma}-w_{\gamma,M}(p)
=
\sum_{m=M}^\infty\frac{(\gamma)_m}{m!}(1-p)^m.
\end{align*}
For large $m$,
\begin{align*}
\frac{(\gamma)_m}{m!}\sim\frac{m^{\gamma-1}}{\Gamma(\gamma)},\ \text{and}\ (1-p)^m\approx e^{-pm}.
\end{align*}
This gives the approximation
\begin{align*}
\frac{R_{\gamma,M}(p)}{p^{-\gamma}}
\approx
Q(\gamma,pM),
\end{align*}
where $Q$ is the regularized upper incomplete gamma function. Requiring relative truncation error at most $\delta$ for $p\ge p_{\min}$ suggests
\begin{align*}
M_{\mathrm{need}}(\gamma;p_{\min},\delta)
:=
\left\lceil\frac{Q^{-1}(\gamma,\delta)}{p_{\min}}\right\rceil.
\end{align*}
A low-$K$ amplification cap $A_{\max}$ gives
\begin{align*}
M_{\mathrm{cap}}(\gamma;A_{\max},N)
:=
\max\{M\le N:\alpha_1^{(\gamma,M,N)}\le A_{\max}\}.
\end{align*}
Using the approximation
\begin{align*}
\alpha_1^{(\gamma,M,N)}
\approx
\frac{M^\gamma}{\Gamma(\gamma+1)N},
\end{align*}
one obtains
\begin{align*}
M_{\mathrm{cap}}(\gamma;A_{\max},N)
\approx
\left\lfloor
\bigl(\Gamma(\gamma+1)A_{\max}N\bigr)^{1/\gamma}
\right\rfloor.
\end{align*}
The feasible window is therefore
\begin{align*}
M_{\mathrm{need}}(\gamma;p_{\min},\delta)
\le
M
\le
M_{\mathrm{cap}}(\gamma;A_{\max},N),\ \text{with}\ M\le N.
\end{align*}
Figure~\ref{fig:frontier} shows a representative frontier for $N=32$.

\bibliographystyle{unsrt}
\bibliography{reference}

@article{maxrl,
  author       = {Tajwar, Fahim and Zeng, Guanning and Zhou, Yueer and Song, Yuda and Arora, Daman and Jiang, Yiding and Schneider, Jeff and Salakhutdinov, Ruslan and Feng, Haiwen and Zanette, Andrea},
  title        = {Maximum Likelihood Reinforcement Learning},
  journal      = {arXiv preprint arXiv:2602.02710},
  year         = {2026},
  eprint       = {2602.02710},
  archivePrefix= {arXiv},
  primaryClass = {cs.LG}
}

@article{objective,
  author       = {Davis, Damek and Recht, Benjamin},
  title        = {What is the Objective of Reasoning with Reinforcement Learning?},
  journal      = {arXiv preprint arXiv:2510.13651},
  year         = {2025},
  eprint       = {2510.13651},
  archivePrefix= {arXiv},
  primaryClass = {cs.LG}
}

@article{williams1992reinforce,
  author       = {Williams, Ronald J.},
  title        = {Simple Statistical Gradient-Following Algorithms for Connectionist Reinforcement Learning},
  journal      = {Machine Learning},
  volume       = {8},
  number       = {3--4},
  pages        = {229--256},
  year         = {1992},
  doi          = {10.1007/BF00992696}
}

@inproceedings{sutton1999policygradient,
  author       = {Sutton, Richard S. and McAllester, David and Singh, Satinder and Mansour, Yishay},
  title        = {Policy Gradient Methods for Reinforcement Learning with Function Approximation},
  booktitle    = {Advances in Neural Information Processing Systems},
  volume       = {12},
  pages        = {1057--1063},
  year         = {1999}
}

@article{schulman2017ppo,
  author       = {Schulman, John and Wolski, Filip and Dhariwal, Prafulla and Radford, Alec and Klimov, Oleg},
  title        = {Proximal Policy Optimization Algorithms},
  journal      = {arXiv preprint arXiv:1707.06347},
  year         = {2017},
  eprint       = {1707.06347},
  archivePrefix= {arXiv},
  primaryClass = {cs.LG}
}

@inproceedings{ahmadian2024rloo,
  author       = {Ahmadian, Arash and Cremer, Chris and Gall{\'e}, Matthias and Fadaee, Marzieh and Kreutzer, Julia and Pietquin, Olivier and {\"U}st{\"u}n, Ahmet and Hooker, Sara},
  title        = {Back to Basics: Revisiting {REINFORCE}-Style Optimization for Learning from Human Feedback in {LLM}s},
  booktitle    = {Proceedings of the 62nd Annual Meeting of the Association for Computational Linguistics},
  pages        = {12248--12267},
  year         = {2024},
  publisher    = {Association for Computational Linguistics}
}

@article{shao2024deepseekmath,
  author       = {Shao, Zhihong and Wang, Peiyi and Zhu, Qihao and Xu, Runxin and Song, Junxiao and Bi, Xiao and Zhang, Haowei and Zhang, Mingchuan and Li, Y. K. and Wu, Y. and Guo, Daya},
  title        = {{DeepSeekMath}: Pushing the Limits of Mathematical Reasoning in Open Language Models},
  journal      = {arXiv preprint arXiv:2402.03300},
  year         = {2024},
  eprint       = {2402.03300},
  archivePrefix= {arXiv},
  primaryClass = {cs.CL}
}

@article{guo2025deepseekr1,
  author       = {Guo, Daya and Yang, Dejian and Zhang, Haowei and Song, Junxiao and Zhang, Ruoyu and Xu, Runxin and Zhu, Qihao and Ma, Shirong and Wang, Peiyi and Bi, Xiao and others},
  title        = {{DeepSeek-R1}: Incentivizing Reasoning Capability in {LLM}s via Reinforcement Learning},
  journal      = {arXiv preprint arXiv:2501.12948},
  year         = {2025},
  eprint       = {2501.12948},
  archivePrefix= {arXiv},
  primaryClass = {cs.CL}
}

@article{yu2025dapo,
  author       = {Yu, Qiying and Zhang, Zheng and Zhu, Ruofei and Yuan, Yufeng and Zuo, Xiaochen and Yue, Yu and Fan, Tiantian and Liu, Gaohong and Liu, Lingjun and Liu, Xin and others},
  title        = {{DAPO}: An Open-Source {LLM} Reinforcement Learning System at Scale},
  journal      = {arXiv preprint arXiv:2503.14476},
  year         = {2025},
  eprint       = {2503.14476},
  archivePrefix= {arXiv},
  primaryClass = {cs.CL}
}

@inproceedings{ouyang2022instructgpt,
  author       = {Ouyang, Long and Wu, Jeff and Jiang, Xu and Almeida, Diogo and Wainwright, Carroll L. and Mishkin, Pamela and Zhang, Chong and Agarwal, Sandhini and Slama, Katarina and Ray, Alex and others},
  title        = {Training Language Models to Follow Instructions with Human Feedback},
  booktitle    = {Advances in Neural Information Processing Systems},
  volume       = {35},
  pages        = {27730--27744},
  year         = {2022}
}

@misc{verl2025,
  author       = {{verl contributors}},
  title        = {{verl}: Volcano Engine Reinforcement Learning for {LLM}s},
  year         = {2025},
  howpublished = {\url{https://github.com/verl-project/verl}},
  note         = {Accessed 2026-05-28}
}

@book{murphy2012,
  author       = {Murphy, Kevin P.},
  title        = {Machine Learning: A Probabilistic Perspective},
  publisher    = {MIT Press},
  year         = {2012}
}

@book{bailey1935,
  author       = {Bailey, W. N.},
  title        = {Generalized Hypergeometric Series},
  series       = {Cambridge Tracts in Mathematics and Mathematical Physics},
  volume       = {32},
  publisher    = {Cambridge University Press},
  year         = {1935}
}

@book{koepf1998,
  author       = {Koepf, Wolfram},
  title        = {Hypergeometric Summation: An Algorithmic Approach to Summation and Special Function Identities},
  publisher    = {Vieweg},
  year         = {1998}
}

\end{document}